# LightNN: Filling the Gap between Conventional Deep Neural Networks and Binarized Networks

Ruizhou Ding, Zeye Liu, Rongye Shi, Diana Marculescu, and R. D. (Shawn) Blanton
Department of Electrical and Computer Engineering
Carnegie Mellon University, Pittsburgh, PA, U.S.A.
{rding, zeyel, rongyes, dianam, rblanton}@andrew.cmu.edu

## ABSTRACT
Application-specific integrated circuit (ASIC) implementations for Deep Neural Networks (DNNs) have been adopted in many systems because of their higher classification speed. However, although they may be characterized by better accuracy, larger DNNs require significant energy and area, thereby limiting their wide adoption. The energy consumption of DNNs is driven by both memory accesses and computation. Binarized Neural Networks (BNNs), as a trade-off between accuracy and energy consumption, can achieve great energy reduction, and have good accuracy for large DNNs due to its regularization effect. However, BNNs show poor accuracy when a smaller DNN configuration is adopted. In this paper, we propose a new DNN model, LightNN, which replaces the multiplications to one shift or a constrained number of shifts and adds. For a fixed DNN configuration, LightNNs have better accuracy at a slight energy increase than BNNs, yet are more energy efficient with only slightly less accuracy than conventional DNNs. Therefore, LightNNs provide more options for hardware designers to make trade-offs between accuracy and energy. Moreover, for large DNN configurations, LightNNs have a regularization effect, making them better in accuracy than conventional DNNs. These conclusions are verified by experiment using the MNIST and CIFAR-10 datasets for different DNN configurations.

## 1. INTRODUCTION
As a widely implemented machine learning model, deep neural networks (DNNs) have shown power in classification, due to their non-linear characteristics, flexible configurations and self-adaptive features [1]. Increasingly in recent years, research has focused on DNNs implemented directly in hardware for a variety of reasons stemming from design requirements or application characteristics. First, real-time classification applications (such as Siri and Google glass [2]) have great sensitivity to latency, thereby making pure hardware implementations better candidates than conventional architectures based on CPUs or GPUs. Second, neural networks implemented in custom ASICs require mostly logic with little complicated control, lending themselves to lower design effort. Third, heterogeneous architectures as a whole appear to be more suitable for DNN implementation due to the combined benefits of CPU, GPU, FPGA and ASIC-based hardware acceleration [3] [4]. In heterogeneous systems, ASICs can handle specific tasks that are required frequently, such as classification tasks in a real-time image recognition application, while CPUs and GPUs can perform the online training. However, hardware implementations of neural networks face difficulties for large scale deployment. With the increased use of DNNs, the number of layers and neurons increases significantly [5]. Google's AlphaGo adopts a 13-layer architecture, with hundreds of filters per layer [6]. Microsoft implements a 152-layer DNN for image classification [7]. An increasing number of neurons and connections in DNNs require significant energy and power, thereby limiting their wide adoption [5].

## 2. RELATED WORK
The two main factors limiting the energy efficiency of computing systems in general, and DNN architectures in particular, are memory accesses and computation. Much of the prior research has focused on the computation. Du *et al.* adopt inexact circuits for multipliers and adders with less dynamic energy consumption [3]. Li *et al.* reduce energy with memristor-based approximators [8], while Kim *et al.* propose an approximate adder to reduce energy [9]. Sarwar *et al.* use alphabet set multipliers that implement multiplication by selection, shifts and adds, reducing energy of neural networks with a pre-computed bank [10]. These works can reduce the energy consumption of DNNs, and some [3] [8] [9] are also compatible with models introduced in this paper.

There has also been significant work on reducing energy of memory accesses. Gupta *et al.* reduce accesses to memory by limiting parameter precision [11]. Furthermore, Venkataramani *et al.* propose an energy-efficient neural network design approach by reducing bit precision of resilient neurons [12], while Zhang *et al.* combine computation approximation and memory access reduction, and propose an approximate computing framework of NNs [13]. Han *et al.* compressed DNNs to reduce weight storage [14]. Hubara *et al.* proposed BNNs, which constrain the weights and activations to be binary values (+1 or -1) and achieve almost no accuracy loss for MNIST and CIFAR-10 datasets [15]. Moreover, the reported accuracy for MNIST is even better than the conventional DNN architecture, especially when the DNN configuration is very large, with many more layers and neurons than typical. On the other hand, BNNs have notably worse accuracy when the smaller Caffe example configurations for MNIST and CIFAR-10 are implemented.

In this paper, we focus on both computation and memory access energy reduction. We propose a new DNN model, LightNN, by replacing multipliers with one shift or a limited number of shifts and adds. Marchesi *et al.* proposed the idea of replacing multipliers of multi-layer perceptrons (MLP) with shifts [16]. Different from their proposed training algorithm where the weights are constrained to be power of 2 after every iteration, we maintain a set of continuous weights and only use the constrained weights in the forward pass. In addition, we test LightNNs on real datasets, and compare their accuracy, energy and area with conventional DNNs and BNNs. LightNNs are also different from Sarwar *et al.* where the weights are broken down into two parts with an equal number of shift operations [10]. Instead, LightNNs maintain weights as a whole and only constrain the total number of shifts. Furthermore, we explore the energy consumption of memory accesses. To the best of our knowledge, our work makes the following contributions:

1. Similar to prior work [3] [8]-[15], we consider the scenario where DNNs are trained in software and used for classification in hardware. Different from prior work, we propose a new neural network architecture, LightNN, which features simpler, more energy-efficient logic. Note that LightNN is *compatible* with prior work [3] [11] [14] since it can be adopted either standalone, or in conjunction with other approaches.
2. Compared to BNNs where all the weights are constrained to $+1$ or $-1$, we relax this constraint and achieve better accuracy than BNNs while still maintaining lower power and area than conventional DNNs. The advantage of this relaxation is that hardware designers can have more options to select the model according to their resource constraint. For a fixed DNN configuration, LightNNs have better accuracy at a slight energy increase than BNNs, yet are more energy efficient with only slightly less accuracy than DNNs. It is worth noting that both LightNNs and BNNs have a regularization effect, thereby achieving higher accuracy than conventional DNNs for large configurations.
3. We compare the accuracy, energy and area of conventional DNNs, BNNs and LightNNs via industrial-strength design simulation, for the MNIST and CIFAR-10 datasets. A set of guidelines for model selection *w.r.t.* accuracy and energy is provided. We also implement a non-pipeline version of conventional DNNs and LightNNs for datasets from the UCI machine learning repository [17], and confirm that LightNNs are compatible with approaches that use limited bit precision for inputs and intermediate results [11].

The rest of this paper is organized as follows. In Section 3, we first introduce conventional DNNs and BNNs, and propose LightNN with its training scheme. Experiment results are shown in Section 4, including accuracy, storage, energy and area for conventional DNNs, BNNs and LightNNs. Section 5 concludes the paper.

## 3. DNN ARCHITECTURE
In this section, we first describe the operation of conventional DNNs. Then, we introduce the architecture and the training scheme of BNNs. Finally, we propose the new DNN architecture, LightNN, and describe its associated training scheme.

### 3.1 Conventional DNNs
An artificial neural network (ANN) is called a DNN when there are more than four layers, including the input and output layers. In the *training phase* with a back-propagation algorithm [18], the loss function, such as the $l_2$-*norm*, *cross-entropy loss,* or *hinge loss*, is computed using output values and data labels, to update the weights used for the linear combination described above. In the *deployment (testing) phase*, the output neuron with the largest value indicates the prediction result. During deployment, the vast majority of computation resources are used for the multiplication within DNNs [3]. Therefore, prior research has focused on replacing multiplication with other types of logic operations that are more energy-efficient. One successful example is BNN.

### 3.2 BNNs
Two types of binarized neural networks (BNNs) have been proposed by Courbariaux *et al*. BinaryConnect [19], a type of BNN, only constrains the weights to $+1$ or $-1$, but leaves the inputs and intermediate results as floating point values. On the other hand, a second BNN known as BinaryNet [15], constrains both weights and intermediate results (activations) to $+1$ or $-1$, and only keeps the input values as floating point.

To train a BinaryConnect or BinaryNet, the classic back-propagation algorithm is adopted. The only change is that during each forward pass, the weights are copied and binarized. Then, the binarized copy of weights is used to compute the gradients. Note that the updated weights are always stored as original floating point values and only in the forward pass of training phase and during testing phase, the weights are binarized [19].

During the testing, BinaryNet is different from BinaryConnect only in that it uses a binarized activation function, thereby having binarized intermediate results. In the testing phase, a sign function $f(x) = sign(x)$ is used as the activation function. In the training phase, hard tanh function, defined as $Htanh(x) = clip(x, -1, 1)$, is used as a substitute for the sign function [15]. The benefit of BinaryNet is that it can further regularize the DNN and can use an XNOR operation to replace multiplications in a hardware implementation.

### 3.3 LightNNs
Replacing multiplications with a shift or a limited number of shifts and adds can serve as a way to build more energy-efficient DNNs and regularize them, thereby producing better accuracy. The detailed architecture and training algorithm are introduced next.

#### 3.3.1 Model Architecture
In binary representations, any parameter $w$ can be written as a sum of powers of two $w = sign(w) \cdot (2^{n_1} + 2^{n_2} + \cdots + 2^{n_K})$, where $K$ is the number of 1s in $w$'s binary representation. A multiplication of two values $w$ and $x$ is equal to several shifts and additions:
$$w \cdot x = sign(w) \cdot (2^{n_1} + 2^{n_2} + \cdots + 2^{n_K}) \cdot x$$
$$= sign(w) \cdot (x \ll n_1 + x \ll n_2 + \cdots + x \ll n_K) \quad (1)$$
where "$x \ll n_1$" means left-shifting $x$ by $n_1$ bits. For negative values of $n_1$, right-shifts are used instead. Assuming $n_1 > n_2 > \cdots > n_K$, smaller $n$ values correspond to a less significant part of the result $w \cdot x$. Furthermore, logical shift units are more energy efficient than multipliers. Therefore, the computation energy consumption can be reduced by converting multiplications to approximate versions using a limited number of shifts (and adds). LightNNs change the computation logic of each neuron. A *k-ones approximation* drops the least significant powers of two in equation (1) such that the resulting value has at most $k$ ones in its binary representation. Figure 1 illustrates a basic example that utilizes a neuron with two inputs. Two weights $w_1$ and $w_2$ are both converted to a 2-ones approximation: $w_1 \approx 2^{n_{11}} + 2^{n_{12}}$, $w_2 \approx 2^{n_{21}} + 2^{n_{22}}$. Therefore, a multiplication $w \cdot x$ is changed to two shifts and one addition. Moreover, when $k = 1$, the equivalent multiplier unit is only a shift.

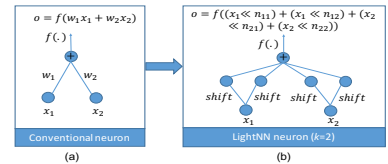

**Figure 1. (a) A conventional neuron and, (b) LightNN neuron implemented using a 2-ones approximation**

Furthermore, we use a stochastic rounding scheme [11]. As opposed to a *rounding-to-nearest* scheme, the *stochastic rounding* scheme finds both the nearest higher value $w_h$ and the nearest lower value $w_l$, and stochastically rounds $w$ to one of them based on a probability distribution:
$$w = \begin{cases} w_h, & \text{with prob } p \\ w_l, & \text{with prob } 1 - p \end{cases}$$

where $p = \frac{w-w_l}{w_h-w_l}$. Intuitively, stochastic rounding ensures that the expected error introduced by the rounding scheme is zero.

### 3.3.2 LightNN Training

We adopt a training algorithm similar to that of BNNs. As shown in Algorithm 1, during each training epoch, the forward pass constrains the weights, and provides intermediate results and loss function. Then, the backward pass computes the derivatives of the loss function over the parameters. After that, the parameters are updated based on the derivatives. It is worth noting that weight constraint occurs only in the forward pass. When updating the weights, we use $w_{t-1} - \eta \frac{\partial l}{\partial w_c}$ instead of $w_c - \eta \frac{\partial l}{\partial w_c}$, where $w_{t-1}$ is the real-value weights after $(t-1)$-th iterations, $w_c$ is the constrained $w_{t-1}$, $\eta$ is the learning rate, and $l$ is the loss function. Therefore, the weights are always accumulated in a floating point form. As stated by Courbariaux *et al.* [19], the reason to maintain high resolution for the weights is that the noise needs to be averaged out by the stochastic gradient contributions accumulated in each weight.

After a LightNN is trained, the last step is to perform the $k$-ones approximation for all weights. Then, the constrained weights are used for testing.

---
**Algorithm 1.** LightNN Training Epoch

**Input:** Training dataset $(x, y)$, where $x$ is input and $y$ is label; parameters after the $(t-1)$-th iteration: $w_{t-1}$ (weights) and $b_{t-1}$ (biases); DNN forward computation function $f(x, w, b)$; $k$ value used for $k$-ones approximation $approx_k(\cdot)$; learning rate $\eta$.
**Output:** Updated weights $w_t$ and biases $b_t$
**For** each mini-batch of $(x, y)$, **do**
1. **Constrain weights:** $w_c = approx_k(w_{t-1})$
2. **Forward:** compute intermediate results and loss function $l$ with $f(\cdot)$, $w_c$, $b_{t-1}$, and mini-batch of $x$
3. **Backward:** compute derivatives $\frac{\partial l}{\partial w_c}$ and $\frac{\partial l}{\partial b_{t-1}}$
4. **Update parameters:** $w_t = w_{t-1} - \eta \frac{\partial l}{\partial w_c}$, and $b_t = b_{t-1} - \eta \frac{\partial l}{\partial b_{t-1}}$

**End For**

---

### 3.3.3 LightNN with Binarized Activations

We can also use a binarized activation function described in Section 3.2 for LightNNs. The advantages of doing so are twofold. First, the multiplication of a weight and an input of a neuron will be limited to $\pm 1$ multiplied by a power of two if the weights are constrained to $k$-ones approximations where $k = 1$. This leads to an increased likelihood of achieving energy reduction within a hardware implementation. Second, binarized activations also inherently perform a regularization, beneficial to eliminating overfitting for large DNN configurations.

## 4. EXPERIMENTS

In this section, we compare conventional DNNs, LightNNs, and BNNs in terms of accuracy, storage, energy consumption and area. We also present a guideline for selecting models for hardware implementations to meet the needs of different applications.

## 4.1 Accuracy

We first introduce the experiment set-up, including the models, datasets, DNN configurations and training approaches. Then, the accuracy results of different models are reported and compared.

### 4.1.1 Set-up

**Models.** Seven models are considered – conventional DNN, LightNN-2, LightNN-1, BinaryConnect, LightNN-2-bin, LightNN-1-bin, and BinaryNet. Table 1 describes their main characteristics. ReLU activation function is adopted in the first four models, while the last three models use the hard tanh function for training and the sign function for testing.

**Datasets.** We test the seven models on two datasets – MNIST and CIFAR-10. The MNIST dataset contains 70,000 gray-scale hand-written images, while CIFAR-10 contains 60,000 colored images for animals or vehicles.

**DNN configurations.** Both multi-layer perceptrons (MLPs) and convolutional neural networks (CNNs) are adopted for MNIST and CIFAR-10. We selected five configurations as shown in Table 2. The basic idea is to include both small and large DNN configurations, to determine how different models perform under varying configurations. **3-hidden** for MNIST and **6-conv** for CIFAR-10 are two large configurations used by Courbariaux *et al.* [15] [19]. **2-conv** for MNIST and **3-conv** for CIFAR-10 are two smaller configurations borrowed from Caffe examples [20]. **1-hidden** for MNSIT is a simple configuration adopted by prior research [10].

**Training approach.** The training algorithm is described in Section 3. Batch normalization and dropout techniques are adopted to accelerate training and avoid overfitting, respectively. The dataset is divided into training set, validation set, and test set. The validation set is used for selecting the best epoch. The same number of total training epochs is applied to all models and the test error of the epoch with the lowest validation error is reported. These models are trained on Theano platform [21]. We use existing open source models [22] to train conventional DNN, BinaryConnect and BinaryNet. Finally, hinge loss function and ADAM learning rule are used to train all seven models [23].

**Table 1. Constraints on seven models.**

| Model | Weights | Activation function | Intermediate results | Inputs |
|---|---|---|---|---|
| Conventional DNN | floating | ReLU | floating | floating |
| LightNN-2 | $\pm(2^{-m_1} + 2^{-m_2})$, $m_1, m_2 = 0,1,\dots,7$ | ReLU | floating | floating |
| LightNN-1 | $\pm 2^{-m}, m = 0,1,\dots,7$ | ReLU | floating | floating |
| BinaryConnect | +1 or -1 | ReLU | floating | floating |
| LightNN-2-bin | $\pm(2^{-m_1} + 2^{-m_2})$, $m_1, m_2 = 0,1,\dots,7$ | Sign | +1 or -1 | floating |
| LightNN-1-bin | $\pm 2^{-m}, m = 0,1,\dots,7$ | Sign | +1 or -1 | floating |
| BinaryNet | +1 or -1 | Sign | +1 or -1 | floating |

**Table 2. Five configurations for two datasets.**

| Dataset | Configuration | Detail |
|---|---|---|
| MNIST | 1-hidden | One hidden layer with 100 neurons |
| | 2-conv | Two convolution layers and two fully-connected layers |
| | 3-hidden | Three hidden layers each with 4096 neurons |
| CIFAR-10 | 3-conv | Three convolution layers and one fully-connected layer |
| | 6-conv | Six convolution layers and three fully-connected layers |

### 4.1.2 Results

**Comparison.** The test error for all configurations and datasets is reported in Table 3. For most configurations (except a few), the accuracy decreases from: conventional, LightNN-2, LightNN-1, LightNN-2-bin, LightNN-1-bin, BinaryConnect, BinaryNet. This is because when we constrain the weights and activations, the model suffers from varying levels of accuracy loss.

**Regularization.** Interestingly, we note that for CIFAR-10 **6-conv** configuration, the conventional DNN performs no better than other models. This shows the regularization effect of weight constraints. When the DNN is relatively large, it tends to overfit and trap within a local optimum. In this case, weight constraints serve as a regularization method to avoid overfitting [19].

**Table 3. Test error and the number of parameters for all models.**

|  |  | MNIST | | | CIFAR-10 | |
| --- | --- | --- | --- | --- | --- | --- |
|  |  | 1-hidden | 2-conv | 3-hidden | 3-conv | 6-conv |
| Number of parameters | | 79,510 | 431,080 | 36,818,954 | 82,208 | 39,191,690 |
| Test error | Conventional | 1.72% | 0.86% | 0.75% | 21.16% | 10.94% |
|  | LightNN-2 | 1.86% | 1.29% | 0.83% | 24.62% | 8.84% |
|  | LightNN-1 | 2.09% | 2.31% | 0.89% | 26.11% | 8.79% |
|  | BinaryConnect | 4.10% | 4.63% | 1.29% | 43.22% | 9.90% |
|  | LightNN-2-bin | 2.94% | 1.67% | 0.89% | 32.58% | 10.12% |
|  | LightNN-1-bin | 3.10% | 1.86% | 0.94% | 36.56% | 9.05% |
|  | BinaryNet | 6.79% | 3.16% | 0.96% | 73.82% | 11.40% |

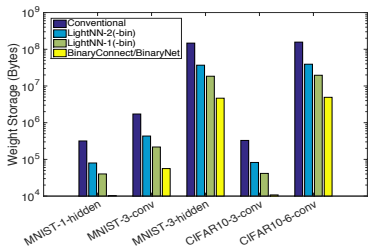

**Figure 2. Storage (for weights) required by different models under varying datasets and configurations.**

## 4.2 Storage requirements

The weight storage requirements for different models are compared in Figure 2. Since the constraint on activations does not affect weight storage, we only show four models. While it has been shown that limited bit precision can also lead to good accuracy [11], we retain the 32-bit representation for the conventional DNN as baseline. Therefore, a weight in conventional DNNs has four bytes, while in BinaryConnect and BinaryNet, it has only one bit. To store a weight $w = \pm 2^{-m}$, LightNN-1 and LightNN-1-bin need four bits: one bit for $sign(w)$ and another three bits for $|m|$. LightNN-2 and LightNN-2-bin need seven bits for a weight $w = \pm(2^{-m_1} + 2^{-m_2})$: one bit for $sign(w)$, three bits for $|m_1|$ and three bits for $|m_2|$. For easier hardware implementation, one byte is used for a weight of LightNN-2 or LightNN-2-bin. Storage affects the number of memory accesses, and is thereby essential to energy consumption, which is shown in Section 4.3.

## 4.3 Energy and Area

### 4.3.1 Set-up

For all seven models under consideration, we designed pipelined implementations with one stage per neuron. The weights and inputs are initially stored in the memory, and fetched to the pipeline stage logic to compute the output for all neurons in that layer; intermediate results are written back. A 65nm commercial standard library is adopted. The logic computations circuit of one neuron is composed using Synopsys Designware commercial IP [24] (e.g. floating point multiplication and addition). The Synopsys Design Compiler [25] is used to generate the gate-level netlist and measure the circuit area. The power consumption of one neuron circuit is calculated using Synopsys Primetime [26]. Cacti [27] is used to obtain the power of memory accesses and registers. While prior work describes approaches (such as data reuse) to optimize the CNN hardware implementation [28], we keep all the models implemented an unoptimized fashion because our main objective is to compare how constraining weights impacts both computation and memory access energy. The size of the register files is chosen to accommodate the data size required for the computation of the largest neuron.

### 4.3.2 Multipliers vs. equivalent multiply units

Power and area of each multiplier or equivalent multiply unit in all models under consideration are explored first. For BinaryConnect and BinaryNet, a multiply unit is simply an XNOR gate [15]. For LightNN-1 and LightNN-1-bin, it is a shift unit. Since operands (e.g., unbinarized weights, activations and inputs) are represented as single-precision float-point., the shift operation is equivalent to an integer addition for the exponent. LightNN-2 and LightNN-2-bin both rely on two shifts and an add operation. The adder required by LightNN-2 is floating point, while LightNN-2-bin only needs an integer adder to perform fixed point addition. The area and power are reported in Figure 3.

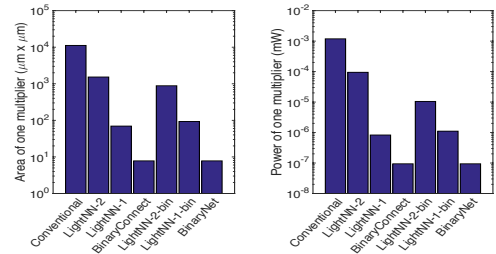

**Figure 3. Area and power of an equivalent multiply unit across all models.**

### 4.3.3 Energy comparison

Figure 4 shows the comparison of energy consumption for all seven models considered. Under the same DNN configuration, conventional DNNs and BinaryNet are always the most and least energy-consuming model, respectively. Furthermore, LightNN-2 is more energy-consuming than LightNN-1 and LightNN-2-bin, both of which consume more energy than LightNN-1-bin. When comparing LightNN-1 and LightNN-2-bin, the former has fewer bits for each weight, and therefore consumes less energy for each memory access, while the latter has more energy-efficient logic. The results in Figure 4 show that LightNN-1 has higher energy consumption than LightNN-2-bin in all configurations except MNIST **1-hidden**. The same comparison holds for the BinaryConnect and LightNN-1-bin, where BinaryConnect has more energy-consuming logic circuitry (e.g., floating point adder) while LightNN-1-bin has larger weight storage.

Although BinaryNet always has the lowest energy consumption under the same configuration, its high accuracy only occurs when the configuration is very large. For example, conventional DNN with **2-conv** can surpass BinaryNet with **3-hidden** in terms of both accuracy and energy consumption.

To explore the energy composition for each model, we report the results in Figure 5. Specifically, the various components of energy are averaged across different configurations and datasets. For the conventional DNNs with floating point circuitry, the most energy-consuming part is the computational portions, while the majority of energy in LightNN-2-bin, LightNN-1-bin and BinaryNet is consumed by memory accesses, though the absolute values are still smaller than that of conventional DNNs. We also break down the logic energy into leakage, switch and internal energy, where the switch energy is caused by switched load capacitance, and the internal energy is due to internal device switching. The leakage, switch and internal energy take 31.6%, 35.2% and 33.2% respectively, averaged on all models and configurations.

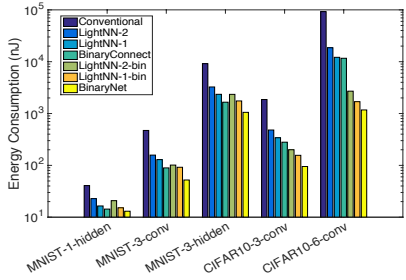

**Figure 4. Comparison of energy consumption of different models under varying datasets/configurations. Energy consumption is measured for inferring a single image.**

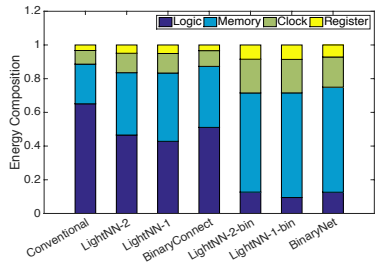

**Figure 5. Energy breakdown for all models, averaged across different datasets and configurations.**

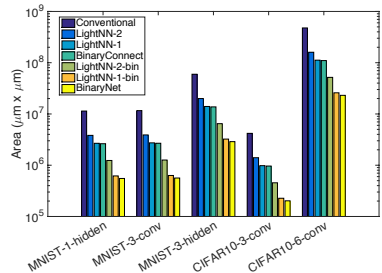

**Figure 6. Comparison of area of different models under varying datasets and configurations.**

### 4.3.4 Area comparison

We also compare the area of the seven models under varying configurations in Figure 6. Note that the reported area includes both logic circuits and register files. Also note that the DNN inference is implemented in a pipeline fashion, and the logic is set to handle the largest neuron count in each configuration. Since more computation modules indicate larger area, but fewer memory fetches, the absolute values for the area encompass the energy consumption reported in Figure 4. However, the comparison of different models within a configuration is still meaningful since they use the same (largest) neuron count. The models follow a consistent order (from larger area to smaller): Conventional DNNs, LightNN-2, LightNN-1, BinaryConnect, LightNN-2-bin, LightNN-1-bin, and BinaryNet.

### 4.3.5 Guidelines for model selection

An interesting question is whether there is a model that always surpasses another in terms of both accuracy and energy, under varying configurations and datasets? The answer is no. However, with constraints on accuracy or energy, some models are more preferable than others. Suppose one will implement DNN for MNIST on hardware with the constraint that energy consumption per inference is below 200nJ. In this case, the LightNN-2 model with **2-conv** configuration is selected, since it has the highest accuracy.

Does higher accuracy always require higher energy? Again, the answer is no. A comparison of different models and configurations are presented in Figure 7. Only the red triangles are the Pareto-optimal ones in terms of accuracy and energy.

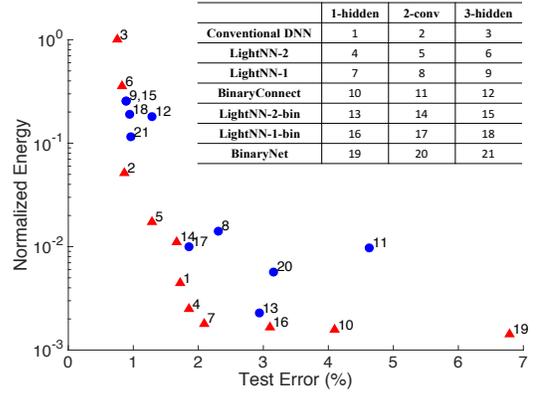

(a)

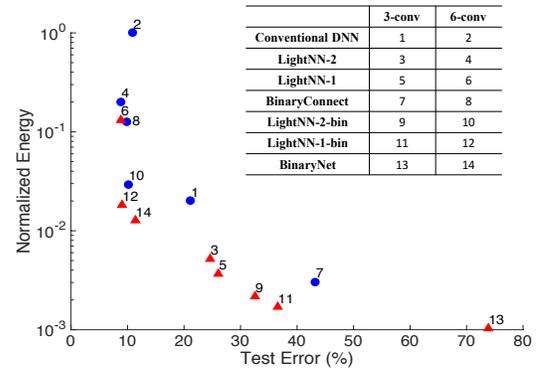

(b)

**Figure 7. Normalized energy and test error of varying models and configurations for (a) MNIST and (b) Cifar-10. Red triangles are Pareto-optimal.**

## 4.4 Non-pipeline Implementation

To further explore how LightNNs perform in a non-pipeline implementation, we implement the ANNs for five benchmarks from the UCI machine learning repository [17]: *abalone, banknote-authentication, transfusion, sinknonsink,* and *balance-scale.* They are chosen because of their small ANN configurations, making direct implementations of whole ANNs doable. The size of the datasets varies from 600 (*balance*) to 200,000 (*sinknonsink*), so a greater coverage is ensured. Instead of computing each neuron at one stage, the non-pipeline implementation builds the whole ANNs for each dataset. Moreover, to confirm that LightNNs are compatible with the use of limited bit precision for inputs and intermediate results [11], we use both 32-bit and 12-bit implementations. Similar to Section 4.3, the ANNs are implemented using Synopsys Designware commercial IP [24].

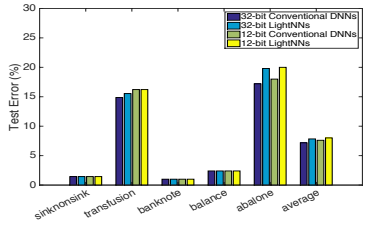
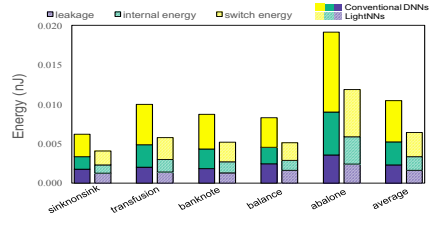
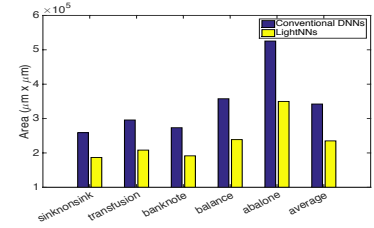

**Figure 8. Test accuracy of conventional DNNs and LightNNs with 32-bit and 12-bit implementation.**

**Figure 9. Energy of 12-bit conventional DNNs and LightNNs.**

**Figure 10. Area of 12-bit conventional DNNs and LightNNs.**

### 4.4.1 Accuracy

Figure 8 shows the accuracy for the testing phase for the five benchmarks. All of them are trained with 2-ones stochastic approximation. For each benchmark, the accuracy of the 12-bit and 32-bit configurations are shown for both the conventional DNNs and LightNNs. From an accuracy standpoint, LightNNs lose only 0.66% and 0.40% accuracy when compared to conventional DNNs for 32- and 12-bit implementations, respectively. Furthermore, the accuracy results for 32- and 12-bit data confirm that the additional error incurred by limiting numerical precision is quite small [11]. Finally, the small accuracy differences between 32-bit and 12-bit LightNNs prove that they have high tolerance for limited precision, thereby showing their compatibility with prior work [11]-[13].

### 4.4.2 Energy and area

12-bit conventional DNNs and LightNNs for the five benchmarks are implemented and all three types of energy consumption are measured: leakage, internal, and switch energy, all shown in Figure 9. The area is also compared in Figure 10. Note that the energy and area are reported only for the logic. The use of LightNNs reduces total energy by 38.8% on average. Both leakage and dynamic energy are reduced, benefiting from the more energy efficient logic implementation. More precisely, the area of a 12-bit multiplier is reduced by 61.6% by the use of the LightNN multiplier. As a result, fewer transistors are required by LightNNs, leading to less leakage and dynamic energy consumption.

## 5. CONCLUSIONS

The increasing demand of neural networks for real-time classification, and the trend of heterogeneous systems leveraging the high speed of ASICs, accelerate the pace of ASIC implementations for DNNs. Due to the increasing size of DNNs, energy and area requirements have become a very challenging problem. LightNNs modify the computation logic of conventional DNNs by making reasonable approximations, and replace the multipliers with more energy-efficient operators involving only one shift or limited shift-and-add operations. In addition, LightNNs also reduce weight storage, thereby decreasing the memory access energy. Experiment results show that LightNNs fill the gap between conventional DNNs and BNNs in terms of accuracy, storage, energy and area, and provide more options for hardware designers to select DNN models based on their accuracy and resource constraints.